\documentclass[11pt]{article}
\usepackage[margin=1.1in,a4paper]{geometry}
\usepackage{natbib}

\usepackage{algorithm, algorithmicx, algpseudocode}
\usepackage{hyperref}

\usepackage{graphicx}
\usepackage{wrapfig}
\usepackage{enumitem}

\usepackage{amsmath,amsfonts,bm}

\def\eqref#1{equation~\ref{#1}}

\def\1{\bm{1}}

\def\rva{{\mathbf{a}}}

\def\rvh{{\mathbf{h}}}
\def\rvu{{\mathbf{i}}}

\def\rvo{{\mathbf{o}}}

\def\rvu{{\mathbf{u}}}

\def\rvx{{\mathbf{x}}}
\def\rvy{{\mathbf{y}}}
\def\rvz{{\mathbf{z}}}

\DeclareMathAlphabet{\mathsfit}{\encodingdefault}{\sfdefault}{m}{sl}
\SetMathAlphabet{\mathsfit}{bold}{\encodingdefault}{\sfdefault}{bx}{n}

\def\sX{{\mathbb{X}}}
\def\sY{{\mathbb{Y}}}

\newcommand{\Ptest}{\hat{P}_{\rm{test}}}

\newcommand{\E}{\mathbb{E}}

\DeclareMathOperator*{\argmin}{arg\,min}

\usepackage{amsmath}
\usepackage{nomencl}
\makenomenclature

\usepackage{mathrsfs}
\usepackage{amsfonts}
\usepackage{optidef}
\usepackage{color}
\usepackage{xcolor}

\newcommand\independent{\protect\mathpalette{\protect\independenT}{\perp}}
\def\independenT#1#2{\mathrel{\rlap{$#1#2$}\mkern2mu{#1#2}}}

\usepackage{graphicx}
\graphicspath{{./graphics/}}
\usepackage{booktabs}
\usepackage{multirow}

\author{
	Susan Wei\thanks{
		School of Mathematics and Statistics
		University of Melbourne;
		\texttt{susan.wei@unimelb.edu.au}}\;
	\and
	Marc Niethammer\thanks{
		Department of Computer Science
		University of North Carolina at Chapel Hill; 
		\texttt{mn@cs.unc.edu}}
}

\title{The Fairness-Accuracy Pareto Front}
\begin{document}
	
	\maketitle
	
	\begin{abstract}
Algorithmic fairness seeks to identify and correct sources of bias in machine learning algorithms. Confoundingly, ensuring fairness often comes at the cost of accuracy. We provide formal tools in this work for reconciling this fundamental tension in algorithm fairness. Specifically, we put to use the concept of Pareto optimality from multi-objective optimization and seek the fairness-accuracy Pareto front of a neural network classifier. We demonstrate that many existing algorithmic fairness methods are performing the so-called linear scalarization scheme which has severe limitations in recovering Pareto optimal solutions. We instead apply the Chebyshev scalarization scheme which is provably superior theoretically and no more computationally burdensome at recovering Pareto optimal solutions compared to the linear scheme. 
\end{abstract}


	\section{Introduction}

The emerging field of algorithmic fairness is concerned with the understanding and correcting of biases in machine learning algorithms.
The scope of algorithmic fairness research is wide, including such goals as understanding post-hoc the cause of bias in both the data collection and the algorithm, correcting pre-hoc the data collection and algorithm design process to mitigate unfairness, defining measurements of fairness, and establishing regulatory guidelines to govern algorithm deployment. In this work, we highlight an underappreciated challenge to ascertaining algorithmic fairness -- while we would like an algorithm to be both fair and accurate, these two objectives actually conflict with each other. As an extreme example, a completely randomized classifier is also a perfectly fair classifier by almost all measures of fairness; we would, however, be loathe to accept the low accuracy and arbitrariness of randomization. 
Thus, recognizing fairness and accuracy as competing objectives, we put to use the concept of Pareto optimality. Specifically, we apply tools from multi-objective optimization to find the fairness-accuracy Pareto front of a deep neural network classifier.

\begin{figure}[h!]
	\centering
	\includegraphics[width=0.5\linewidth, trim={0 1cm 16cm 3cm},clip]{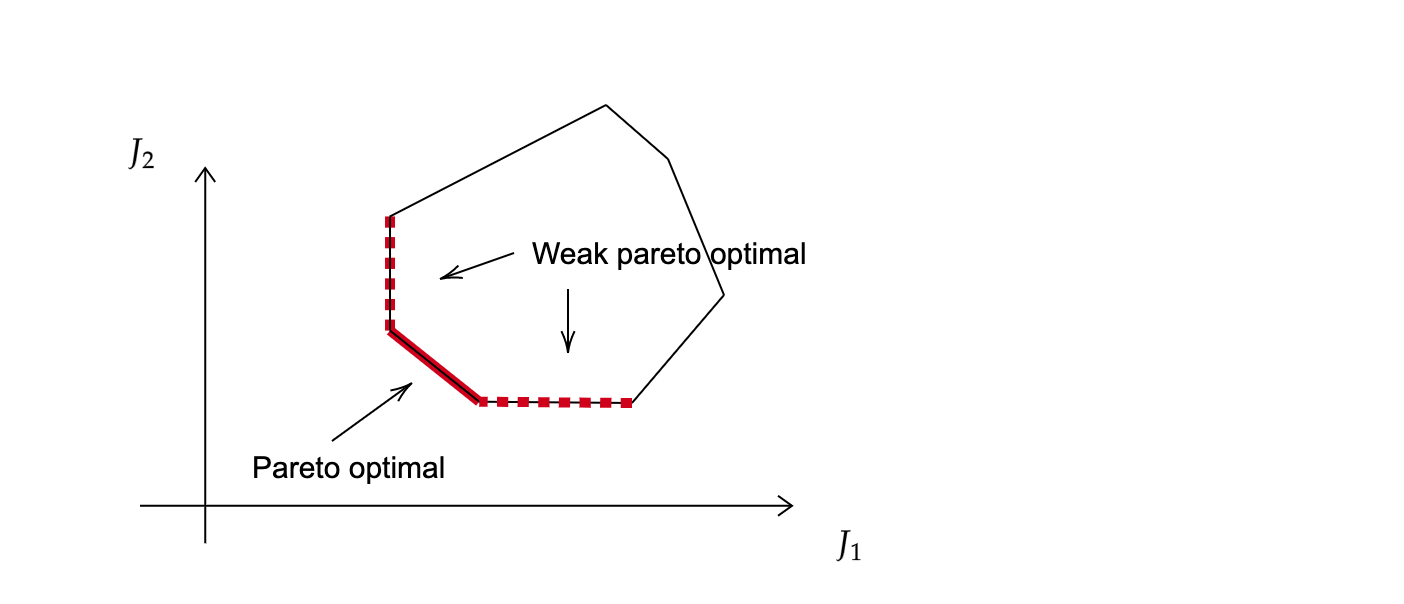}
	\caption{This diagram illustrates the difference between weakly Pareto optimal and Pareto optimal points. All other points in the feasible region, represented by the polygon, are dominated points. The Pareto front itself, consisting of Pareto optimal points, is highlighted in solid red while the weakly Pareto optimal pints that are not Pareto optimal are highlighted in dashed red.}
	\label{fig:pf}
\end{figure}

A typical multi-objective optimization problem is given by $\argmin_{\theta} [J_1(\theta), \ldots, J_k(\theta)]^T$,
where each $J_i$ is a loss function from some parameter space $\Theta$ to $\mathbb R$.
The task is made difficult by the fact that it is rarely possible to find a $\theta^*$ such that all individual objectives are minimized \textit{simultaneously at $\theta^*$}. 
In such circumstances, one can appeal to Pareto optimality, a well-established notion in optimization \citep{emmerich2018tutorial} that is used often in engineering and economics. Pareto optimality is based on the following partial order: for $a,b \in \mathbb R^p$, we say $a \le b$ if and only if every component of $a$ is less than or equal to the corresponding component of $b$. 
We say $\theta \in \Theta$ is \textbf{Pareto optimal} if and only if it is non-dominated, i.e.\ there does not exist any $\tilde \theta \in \Theta$ such that $(J_1(\tilde \theta), \ldots, J_k(\tilde \theta)) \le (J_1(\theta), \ldots, J_k(\theta))$ with at least one strict inequality. 
Similarly, we say $\theta$ is \textbf{weakly Pareto optimal} if there exists no other $\tilde \theta$ such that $J_i(\tilde \theta) < J_i(\theta)$ for all $i=1,\ldots,k$. 
Note that while a Pareto optimal point is also weakly Pareto optimal, the converse is not true. 
The \textbf{Pareto front} is the set of all Pareto optimal points. Figure \ref{fig:pf} displays a toy Pareto front for a bi-objective optimization problem.

Almost all techniques for finding the Pareto front are based on scalarization, i.e.,  turning the vector objective into a scalar one by aggregating the individual objectives. A successful scalarization scheme should satisfy the following two desiderata. 
\begin{enumerate}[label=D\arabic*. , wide=0.5em,  leftmargin=*]
	\item The solutions to the scalarized objective are (weakly) Pareto optimal.
	\item All Pareto optimal points are recoverable from the scalarization scheme.
\end{enumerate}

\begin{figure}[h!]
	\centering
	\includegraphics[width=0.45\linewidth, trim={0 4cm 16cm 3cm},clip]{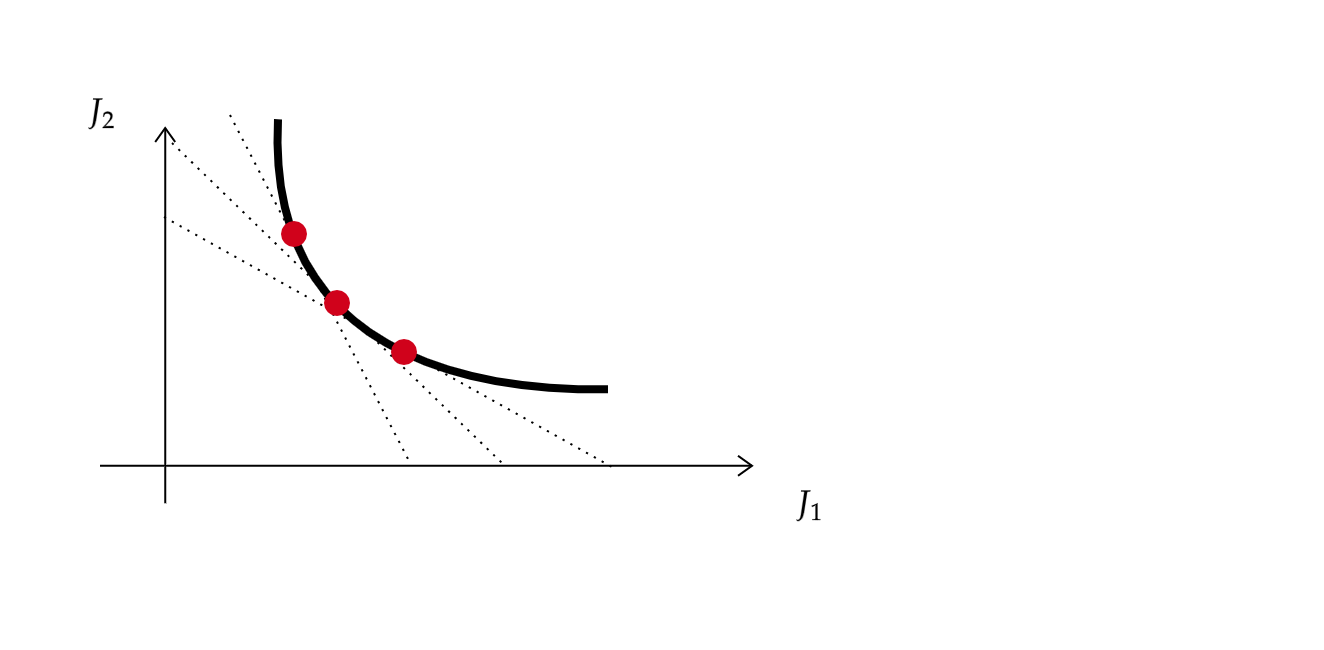}
	\includegraphics[width=0.45\linewidth, trim={0 4cm 16cm 3cm},clip]{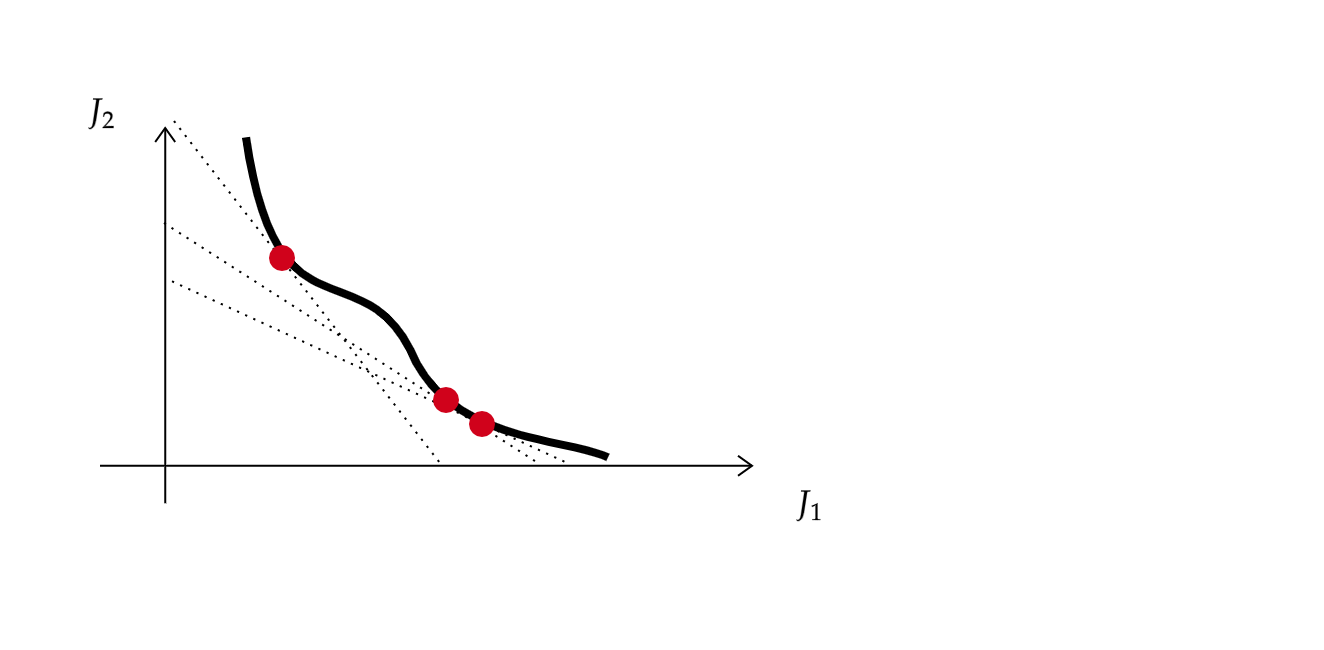}
	\caption{Two hypothetical Pareto fronts are displayed: convex (left) and non-convex (right). The dashed lines represent different weight combinations in the LSP \eqref{eq:convexcombo_general}. Pareto optimality occurs only at the tangential points of the dashed lines. It is clearly that the LSP cannot recover Pareto optimal points that reside in the non-convex region of the Pareto front.}
	\label{fig:lsp}
\end{figure}

\paragraph{Linear scalarization}
The simplest scalarization scheme is the linear scalarization scheme \citep{emmerich2018tutorial}. Here the vector objective is replaced with a weighted sum, which we will call a \textbf{linear scalarization problem (LSP)},
\begin{equation}
	\sum_{i=1}^k \lambda_i J_i, \quad \lambda_i \in \mathbb R_{>0}, \quad i=1,\ldots,k.
	\label{eq:convexcombo_general}
\end{equation}
Sometimes the constraint, $\sum_{i=1}^k \lambda_i = 1$, is imposed, but this is not in fact necessary, e.g., if the individual objectives are not normalized. 
The linear scalarization scheme satisfies desideratum D1. Specifically, no matter the weights $\lambda_i$ chosen, a solution of the linear scalarization problem is weakly Pareto optimal, see Proposition 8 in \citep{emmerich2018tutorial}. However, the linear scalarization scheme critically fails desideratum D2. In fact, it is only able to recover points on the \textit{convex hull} of the Pareto front \citep{Das1997}. 
In particular, when the Pareto front of interest is non-convex, the naive linear scalarization scheme may only find such $\theta$ that strongly favors one of the objectives \citep{emmerich2018tutorial}. 
We illustrate this phenomenon in Figure \ref{fig:lsp}.

Despite the known drawbacks to the linear scalarization scheme for finding Pareto optimal points, almost all train-time algorithmic fairness methods, to be reviewed in the following section, ascribe to this approach. To see this, we note that many popular algorithmic fairness techniques minimize an objective of the form
$$
\text{error}(\theta) + \lambda \times \text{unfairness}(\theta), 
$$ 
where $\theta$ is the model parameters to be learned. 
That is, they regularize the learning algorithm by employing the fairness criterion as a penalty term during training \citep{Hardt2016,Joseph2016,Zafar2017AISTATS,Zafar2017WWW}. 
It is easy to recognize this constrained optimization as the linear scalarization problem in Equation \eqref{eq:convexcombo_general}. Thus existing algorithmic-fairness methods can be expected to inherit all the limitations of the linear scalarization scheme in regards to finding Pareto optimal points. 

\paragraph{Proposal}
In this work, we use instead the Chebyshev scalarization scheme (Section \ref{sec:pareto}) to find the fairness-accuracy Pareto front of a deep neural network classifier. The Chebyshev scheme is no more computationally burdensome than the linear scheme while enjoying theoretically superior performance when the Pareto front is non-convex (as it may well be). 
The final deliverable is a set of neural networks spanning the fairness-accuracy space from the high-accuracy-low-fairness corner to the high-fairness-low-accuracy corner. This could then be consulted by an interested party, e.g.\ a relevant regulatory body, to decide on a particular classifier with the acceptable trade-off.

In recognition of the dominant role inhabited by deep learning in predictive tasks, our examples are limited to deep neural network classifiers. We should note however that nothing prevents the application of the proposed framework to classifiers besides neural networks, as long as the classifier is amenable to training by backpropagation. This includes models that are more familiar to the statistical audience such as logistic regression and support vector machine. In fact, logistic regression is simply a multi-layer perception with no hidden layers and thus a special case of the feedforward neural network we will work with. 

\section{Related work}
\label{sec:relwork}

\paragraph{Multi-objective optimization in machine learning and statistics}
Multi-objective optimization techniques are gaining increasing traction in machine learning \citep{yaochu_jin_pareto-based_2008}. Bayesian optimization \citep{zhang2020random} and reinforcement learning \citep{moffaertMultiObjectiveReinforcementLearning2014} seem to have particularly embraced ideas from multi-objective optimization. The multiple-gradient descent algorithm, a common gradient-based multiobjective optimization technique, has been applied to kernel learning \cite{li2014pareto} and Bayesian optimization \cite{shah2016pareto}.
Though multiple-gradient descent in its original form does not scale up to the high dimensionality of the parameter space of a neural network, \citet{sener2018multi} proposed a workaround that demonstrated good results on standard deep learning benchmarking datasets. Another work in this vein seeks to further articulate user preference for a specific tradeoff, by dividing a deep multi-task learning problem into different subproblems \citep{lin2019pareto}.

\paragraph{Algorithmic fairness}

Let us now briefly review those works in algorithm fairness that are concerned with correcting the bias of machine learning algorithms. These methods are set apart according to the stage in which intervention is taken. The first class of methods attempts to remove bias from the input data itself. These methods rest on the premise that once proper preprocessing is accomplished, any classifier can be used to produce subsequently fair predictions \citep{Kamiran2012,Feldman2015,Calmon2017, Johndrow2019}. 
On the other hand, post-processing techniques directly operate on the classifier output and are, in principle, amenable to any classifier. The technique in \citet{Hardt2016} for instance seeks to learn a monotone transformation of the classifier output to enforce demographic parity or equalized odds, two standard definitions of fairness that we will review in Section \ref{sec:fairness}. 

The third type of algorithmic fairness methods directly intervenes during training. Generally speaking, these train-time methods add the fairness criterion as a regularization term to the main objective of minimizing predictive error \citep{Calders2010,Kamishima2011, Zafar2017AISTATS,Zafar2017WWW,Zafar2017NIPS,Bechavod2017,Agarwal2018,Narasimhan2018}. While early works of this type \citep{Berk2017,Zafar2017WWW,Bechavod2017} focused on simple machine learning algorithms such as logistic regression, more recent work \citep{Beutel2017,Wadsworth2018,Madras2018,ManishaGujar2018} can handle more complex models such as neural networks. 

The proposed work also intervenes at train time. It appears to be the first in the algorithmic fairness literature to advocate for the important role that Pareto optimality should play, considering that fairness and accuracy are competing objectives of interest. Many methods do not consider the trade-off curve or when they do, they do not use proper techniques to recover Pareto optimal points in the fairness-accuracy space.

\section{The fairness-accuracy Pareto front}
\label{sec:pareto}
In this section, we apply the Chebyshev scalarization scheme to estimate the fairness-accuracy Pareto front of a neural network binary classifier.
Suppose the data consists of input variables $\rvx \in \sX \subset \mathbb R^{p}$ standardized to mean zero and unit variance, binary response $\rvy \in \sY$ indicating class membership, and binary sensitive variable $\rva$. Let all discrete variables be dummy encoded.
A feedforward neural network is a repeated composition of affine transformation followed by nonlinear transformation. Let $m_0 = p$ and define $w^{(l)}  \in \mathbb R^{m_l \times m_{l-1}}$ and $b^{(l)}  \in \mathbb R^{m_l}, l = 1, \ldots, L$ as the parameters in the $l$-th layer of a fully-connected feedforward neural network with $L$ layers. Consider the affine transformation $h^{(l)}: \mathbb R^{m_{l-1}} \to \mathbb R^{m_l}$ and nonlinear transformation $v^{(l)}$ given by
\begin{align*}
	h^{(l)} &= w^{(l)}  v^{(l-1)} + b^{(l)}, \quad l = 1, \ldots, L \\
	v^{(l)} &= \sigma^{(l)} \circ h^{(l)}, \quad l = 1, \ldots, L
\end{align*}
where $v^{(0)} = id$ is the identity function and $\sigma^{(l)}: \mathbb R^{m_l} \to \mathbb R^{m_l}$  is a nonlinear transformation (more frequently called the activation function) that acts component-wise. 

The activation function in the final layer, $\sigma^{(L)}$, will be restricted to the sigmoid function, $(1+\exp(-x))^{-1}$, so that the classifier outputs scores between $0$ and $1$. We use the ReLU activation function, $max(0,x)$, in all other layers for our experiments. Let $\rvh^{(l)}_i$ be shorthand for the application of the function $h^{(l)}$ to input feature $\rvx_i$, i.e.\ 
$$\rvh^{(l)}_i = h^{(l)}(\rvx_i).$$
Collecting all parameters $(w^{(l)},b^{(l)}), l=1, \ldots, L$ into the parameter vector $\theta$, let $f_\theta: \sX \to [0,1]$ be a fully-connected feedforward neural network with parameter $\theta \in \Theta$ constructed as above, i.e., the neural network acts as $f_\theta(x) = v^{(L)}(x)$.


Let $P$ be the joint distribution of the data $(\rvx,\rva,\rvy).$
The accuracy of $f_\theta$ will be measured by its risk, 
$$
R(\theta) = \E_{P} \mathcal L(f_\theta(\rvx),\rvy),
$$ 
where $\mathcal L$ is a loss function. Since we are interested in binary classification, we will limit future discussion to the binary cross-entropy loss, $\mathcal L: [0,1] \times \{0,1\} \to \mathbb R$ given by 
$$\mathcal L(\hat p, y) = y \log \hat p +(1-y) \log (1-\hat p).$$
Next, let $U(\theta)$ be some population measure for the fairness of $f_\theta$. We shall take the convention that higher values of $U$ are undesirable, yielding algorithms that are more unfair.
Since we wish for the learning algorithm $f_\theta$ to be both accurate and fair, we have the following \textit{vector} objective function
\begin{equation}
	\argmin_{\theta} \begin{bmatrix}
		R(\theta)\\
		U(\theta)
	\end{bmatrix}.
	\label{bivariate_Err_fair_pop}
\end{equation}



\begin{figure}
	\centering
	\includegraphics[width=0.45\linewidth, trim={0 4cm 16cm 3cm},clip]{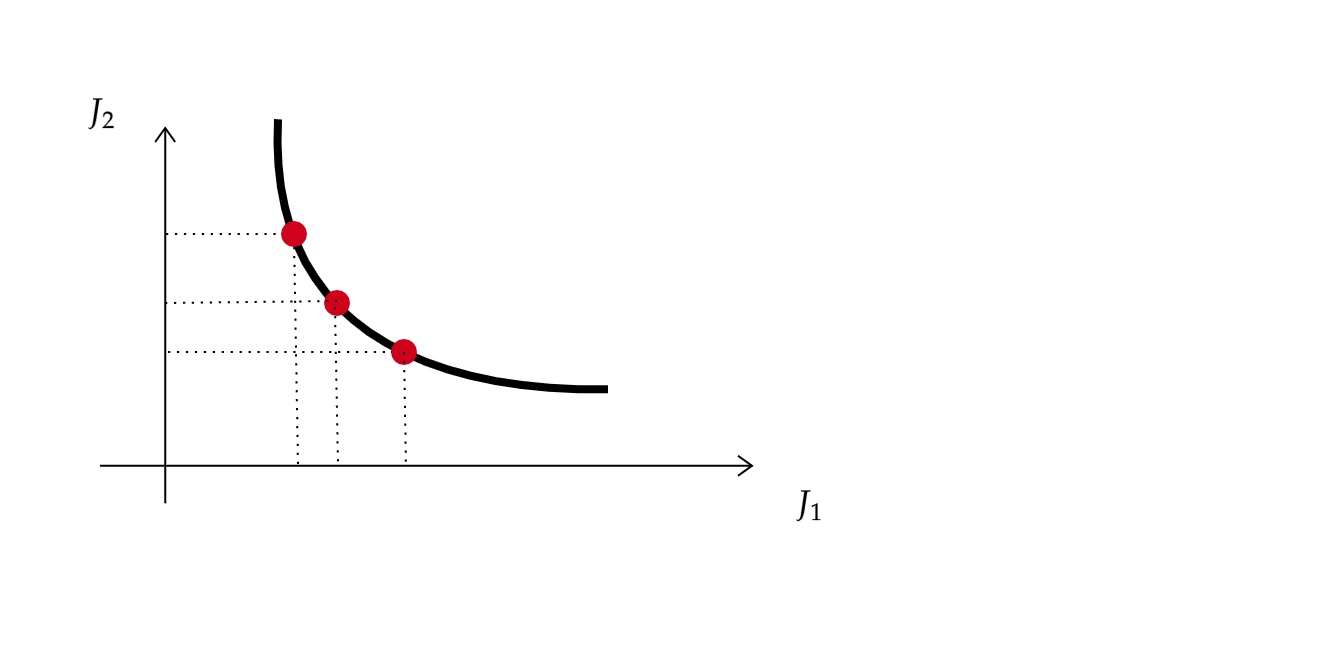}
	\includegraphics[width=0.45\linewidth, trim={0 4cm 16cm 3cm},clip]{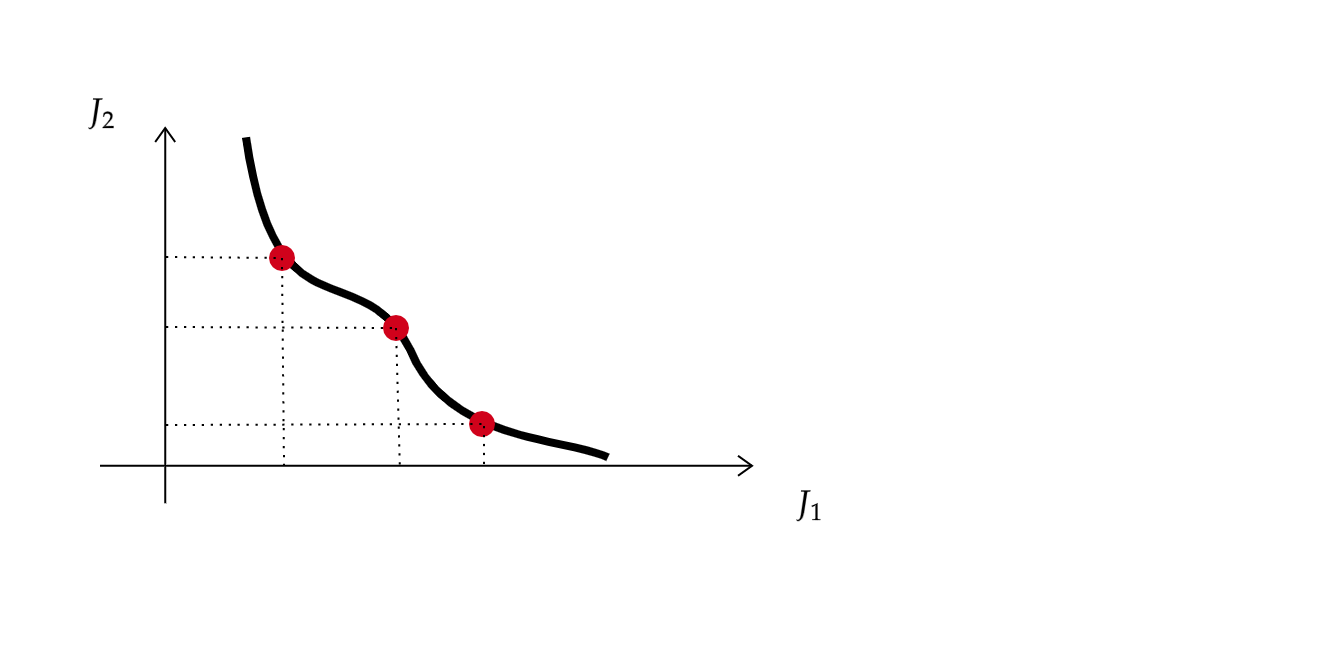}
	\caption{We again consider the two hypothetical Pareto fronts from Figure \ref{fig:lsp}. This time the dashed lines represent different weight combinations in the CSP \eqref{chebyScalarObjective}. We can see that the CSP, in contrast to the LSP, can recover Pareto optimal points that reside in the non-convex region of the Pareto front.}
	\label{fig:csp}
\end{figure}

\paragraph{Chebyshev scalarization}
As an alternative to the linear scalarization scheme in \eqref{eq:convexcombo_general}, we consider the Chebyshev scalarization scheme \citep{Ehrgott2000,Giagkiozis2015} whereby the vector objective, $[J_1(\theta), \ldots, J_k(\theta)]^T$,  is replaced with what we will call the \textbf{Chebyshev scalarization problem (CSP)}, 
\begin{equation}
	\max_{i=1,\ldots, k}  \{\lambda_i |J_i(\theta) - z_i^*|\}, \quad \lambda_i \in \mathbb R_{>0}, \quad i=1,\ldots,k
	\label{chebyScalarObjective}
\end{equation}
where $z_i^* = \inf_\theta J_i(\theta)$.
The Chebyshev scalarization satisfies both desiderata D1 and D2. Namely the solutions to \eqref{chebyScalarObjective} are at least weakly Pareto optimal and hence possibly Pareto optimal but not necessarily. Furthermore, any Pareto optimal solution can be obtained for some configuration of $\lambda_i$'s, see Proposition 10 of \citep{emmerich2018tutorial}. In particular, this means the Chebyshev scheme can find Pareto optimal points that reside in the nonconvex area of the Pareto front in contrast to the linear scalarization scheme. Figure \ref{fig:csp} contains an illustration of this. However, we caution that this does \textit{not} mean that every CSP solution is Pareto optimal because again the solution may be weakly Pareto optimal.

More advanced scalarization schemes than either the LSP or the CSP certainly exist. However, we focus on the CSP in this paper as a first step in applying MOO to algorithmic fairness. CSP is an obvious candidate given its clear superiority to LSP. Furthermore there is no additional computational overhead to solve the CSP compared to the LSP. 


\paragraph{Estimates of the individual components}
Solving \eqref{chebyScalarObjective} first requires estimation of the population unknowns $R(\theta)$ and $U(\theta)$. Since most learning algorithms proceed by empirical risk minimization, we can without loss of generality estimate $R(\theta)$ with the plug-in estimator. Specifically let 
$
R_n(\theta) = \mathbb P_n \mathcal L(f_\theta(\rvx),\rvy)
$
where $\mathbb P_n = \frac{1}{n} \sum_{i=1}^n \delta_{(\rvx_i,\rva_i,\rvy_i)}$ is the empirical measure for the training set $(\rvx_1,\rva_1,\rvy_1), \ldots, (\rvx_n,\rva_n,\rvy_n) \sim P$.  
We will defer discussion of the estimation of $U(\theta)$ to Section \ref{sec:fairness} in the context of specific fairness metrics. For now, let $U_n(\theta) $ denote some estimator of $U(\theta)$ based on the dataset of sample size $n$. (It does not need to be the plug-in estimator.) 

\paragraph{Standardization}
The individual estimators $R_n$ and $U_n$ should be properly scaled so that the weights in \eqref{fig:csp} can better articulate user preferences. We will focus on discussing the standardization of $R_n$ and $U_n$ in the context where $f_\theta$ is a neural network. As mini-batches are almost always employed during training of a neural network, the risk and unfairness measures will be evaluated on mini-batches. Let $\{(\rvx_1^*,\rva_1^*,\rvy_1^*), \ldots, (\rvx_b^*,\rva_b^*,\rvy_b^*)\}$ be a mini-batch sample drawn without replacement from $\mathbb P_n$. Then define the mini-batch estimate of the risk as 
$
R_b^*(\theta) = \frac{1}{b} \sum_{i=1}^b \mathcal L(f_\theta(\rvx_i^*),\rvy_i^*).
$
Similarly, let $U_b^*(\theta)$ be the mini-batch estimate of the fairness measure.

First we solve \eqref{chebyScalarObjective} for $\lambda = 0$. This will not typically be an exact solution. For instance if $f_\theta$ is a neural network, \eqref{chebyScalarObjective} can only be solved approximately, e.g.\ optimization via stochastic gradient descent. Let $R_{min}$ and $R_{max}$ denote, respectively, the minimum and maximum value of $R_b^*$ observed across the mini-batches over all epochs. We can similarly obtain $U_{min}$ and $U_{max}$ by solving \eqref{chebyScalarObjective} for $\lambda = 1$. Then we standardize the accuracy and fairness components as follows
$$
\tilde R_n(\theta) = (R_n(\theta) - R_{min})/(R_{max} - R_{min})
$$ 
and 
$$\tilde U_n(\theta) = (U_n(\theta) - U_{min})/(U_{max} - U_{min}).
$$

\paragraph{Pareto front candidates}
Using the Chebyshev scalarization scheme leads us to Pareto front candidates given by the set $\{\hat \theta_n^\lambda: \lambda \in \Lambda\}$ where 
\begin{equation}
	\hat \theta_n^\lambda = \argmin_\theta \max\{ (1-\lambda) \tilde R_n(\theta), \lambda \tilde U_n(\theta)  \}.
	\label{chebyScalarObjective_empirical}
\end{equation}
and $\Lambda \subset [0,1]$ is a finite set of $\lambda$ values which includes $\lambda=0$ and $\lambda=1$. Note that \eqref{chebyScalarObjective_empirical} follows directly from \eqref{chebyScalarObjective} since the standardization step renders $z_i^* = 0$ for all $i$. Ideally, we would then solve \eqref{chebyScalarObjective_empirical} for a dense $\Lambda$. While intuitive, evenly distributed $\lambda$'s in the interval $[0,1]$ should be avoided as this often produces solutions that form clumps in the Pareto front, i.e.\  evenly distributed $\lambda$'s in $[0,1]$ do \textit{not} produce evenly distributed points in the multi-objective space. The experiments in Section \ref{sec:experiments} employ approximately evenly-spaced values of $\lambda$ on the log scale for the set $\Lambda$. Future work might involve more sophisticated techniques from multi-objective optimization, e.g.\  the Normal-Boundary-Interactive \citep{Das2000} which adaptively selects $\lambda$.

\begin{algorithm}[tb]
	\caption{Fairness-accuracy Pareto front candidates}
	\label{alg}
	\begin{algorithmic}
		\State {\bfseries Input:} multiple splits of the data into training set (sample size $n$) and testing set (sample size $T$), and a finite set $\Lambda \subset [0,1]$ which includes $\lambda=0$ and $\lambda=1$.
		\State Initialize \textsc{candidates} $= \emptyset$.
		\For{each training-testing split}
		\State Solve \eqref{chebyScalarObjective_empirical} on the training set to obtain $\hat \theta_n^\lambda$ for each $\lambda \in \Lambda$
		\State Add to \textsc{candidates} the set $\{\hat \theta_n^\lambda: \lambda \in \Lambda\}$ 
		\EndFor
		\State {\bfseries Output:} \textsc{candidates}
	\end{algorithmic}
\end{algorithm}


Finally, to get a robust estimate of the Pareto front, we should average out, so to speak, the randomness associated with the training-testing split. If $M$ pairs of training and testing splits are considered, then \textsc{candidates}, the output of Algorithm \ref{alg}, contains in total $M |\Lambda|$ Pareto candidates. 

\paragraph{Visualization of results}
To visualize the estimated Pareto front, we plot each element of \textsc{candidates}, the output  of Algorithm \ref{alg}, in the fairness-accuracy space as evaluated on the test set. Specifically, we calculate the out-of-sample risk of $\theta$ as
\begin{align*}
	R(\theta; \Ptest) &= \E_{\Ptest} \mathcal L (f(\rvx;\theta),\rvy) =  \frac{1}{T} \sum_{i=1}^T \mathcal L (f(\rvx_i^*;\theta),\rvy_i^*) 
\end{align*}
where $\Ptest= \frac{1}{T} \sum_{i=1}^T \delta_{(\rvx_i^*,\rva_i^*,\rvy_i^*)}$ is the empirical measure of the test set $\{(\rvx_i^*,\rva_i^*,\rvy_i^*)\}_{i=1}^T$ drawn independently from the (unknown) distribution $P$. Let $U(\theta; \Ptest)$ denote the fairness metric assessed on the test set. Again it need not be a plug-in estimator. Each Pareto front candidate can then be visualized by plotting $R(\theta; \Ptest)$ versus $U(\theta; \Ptest)$ for every $\theta$ in \textsc{candidates}. We further cull these candidates by keeping only the non-dominated points, forming the final Pareto front estimate. 

Each point on the estimated Pareto front corresponds to a different neural network classifier $f_\theta$. A practitioner can examine the estimated Pareto front to select the neural network with an acceptable trade-off between fairness and accuracy.

\section{Fairness measures}
\label{sec:fairness}
We will be examining the fairness-accuracy Pareto front corresponding to a variety of fairness notions. In this section we begin by reviewing two common notions of fairness -- demographic and conditional parity. We then review causal approaches to fairness. Along the way we introduce new estimators of conditional parity and introduce a new causal fairness measure, each of which may be of independent interest.

\subsection{Demographic parity and conditional parity}
Many standard fairness measures involve checking the statistical independence between the prediction and the sensitive attribute. 
The classifier $f_\theta(\rvx)$ is said to exhibit \textbf{demographic parity} with respect to the sensitive attribute $\rva$ if 
$
f_\theta(\rvx) \independent \rva,
$
where $\independent$ stands for independence. 
Intuitively, demographic parity assesses whether the predicted score depends on the sensitive variable. For example, a classifier predicting if a convicted criminal will re-offend exhibits demographic parity with respect to race if the distribution of $f_\theta(\rvx)$ is the same irrespective of race.
Despite its intuitiveness, the drawbacks to demographic parity are well-documented \citep{Hardt2016,Kleinberg2018}. For instance, when the base rates differ across values of the sensitive attribute, satisfying demographic parity can come at the cost of discrimination.

Conditional parity, a term coined in \citet{Ritov2017}, is another notion of fairness that encompasses several fairness measures. The prediction $f_\theta(\rvx)$ is said to exhibit \textbf{conditional parity} with respect to sensitive attribute $\rva$ conditioned on $\rvu$ if 
$
f_\theta(\rvx) \independent \rva \mid  \rvu.
$
Conditional parity in fact unifies several existing fairness definitions. For instance, the notion of equalized odds, introduced in \citet{Hardt2016}, is recovered by setting $\rvu$ to $\rvy$, the true target class membership itself. 

That the notions of demographic and conditional parity can strongly differ and may lead to seemingly paradoxical results was strikingly illustrated in \citet{bickel1975sex} for graduate admissions at UC Berkeley.
Consider a classifier predicting whether an applicant should be admitted to graduate school. One may desire admission decisions to be independent of gender (demographic parity), or independent of gender \textit{conditional} on a particular university department (conditional parity). Depending on which fairness measure is employed, radically different conclusions may be reached.  


We propose to assess demographic and conditional parity using the mean-variance statistic proposed in \citet{Cui2015ModelFreeFS}. The development of these estimators is not central to the proposed work but may be of separate interest so we defer its presentation to Appendix \ref{appendix:mv_condpar}.

\subsection{Causal fairness in the overlap population}
Taking the causal approach to defining fairness means replacing the question ``Is the learning prediction (conditionally) dependent on the sensitive attribute?" with the question ``Does the sensitive attribute have a \textit{causal effect }on the learning predictions?"
Causal approaches to defining fairness \citep{Kusner2017,Kilbertus2017,Loftus2018,Khademi2019} are motivated by the consideration that selection bias will result in a study population different from the target population. This is precisely the situation in many datasets of interest in algorithmic fairness. For instance, in predicting recidivism, the training data suffers selection bias leading to a training population (re-offenders that were caught) likely different from the target population (all would-be re-offenders).

We propose a new causal fairness measure which assesses the average effect of $\rva$ on the predicted probabilities $f_{\theta}(\rvx)$ in the so-called \textit{overlap population} introduced in \citet{LiLock2018}. The overlap population is of particular policy relevance. It is the subpopulation where the sensitive attribute $\rva$ occurs with equal probability, given the prediction inputs $\rvx$. For instance, in the context of recidivism prediction, the overlap population is the subpopulation with similar inputs, e.g.\ criminal record, demographic characteristics, who could easily be either Caucasian or from a minority group. In contrast, many traditional causal estimands focus on unrealistic target populations comprised of individuals who are atypical for their particular value of $\rva$.

Adopting the potential outcome framework of \citet{Imbens2015} and assuming the Stable Unit Treatment Value Assumption, suppose the outcome of interest $\rvo$ takes on one of two potential outcomes, either
$\rvo(0)$ or $\rvo(1)$,
depending on whether $\rva = 0$ or $\rva = 1$. Note that $\rvo =  \rvo(1) \rva +  \rvo(0) (1-\rva) $, i.e.\ we can only ever observe one of the two potential outcomes.
Furthermore, suppose the condition of \textit{unconfoundedness} is satisfied, i.e.\ $\rva$ is independent of $\{\rvo(0),\rvo(1)\}$ conditional on $\rvx$. 


The \textit{average treatment effect for the overlap population} (ATO) introduced in \citet{LiLock2018}
is given by 
\begin{align*}
	&\tau_{ATO}(\rvo; P) = \frac{\E_{P} [ e(\rvx) (1-e(\rvx)) ( \E(\rvo(1) \mid \rvx) - \E(\rvo(0) \mid \rvx) )] }{\E_{ P} [e(\rvx)(1-e(\rvx))]}
\end{align*}
where $e(x) = \displaystyle P(\rva= 1 \mid \rvx = x)$ is also called the propensity score.
Again, the term overlap refers to the fact that the ATO articulates the causal effect among the \textbf{overlap population} which consists of subjects, given their covariates, who could appear with substantial
probability in either value of the sensitive attribute.  
Consider measuring fairness as the causal effect of $\rva$ on the predicted probability $f_\theta(\rvx)$ in the overlap population, i.e.\ 
\begin{align*}
	U(\theta) &= \left \vert \tau_{ATO}(f_\theta(\rvx);P) \right \rvert.
\end{align*}
Its evaluation on the test set is straightforward. Define 
$$
U(\theta,\Ptest) = \left \vert \tau_{ATO}(f_\theta(\rvx);\Ptest) \right \rvert.
$$

Estimation of the ATO causal estimand first requires estimating the propensity score. In our experiments, we used a neural network to do this. We then calibrated the predicted probabilities using the temperature scaling procedure of \citet{Guo2017}. Further details follow in the experiments section. 
For now, let $\hat e(x)$ denote the propensity score estimate.
\citet{LiLock2018} proposed estimating $\tau_{ATO}$ as follows:
\begin{equation}
	\hat \tau_{ATO}(\rvo) = 
	\frac{\sum_{i=1}^n \rva_i \rvo_i w_i}{\sum_{i=1}^n \rva_i w_i} -  \frac{\sum_{i=1}^n (1-\rva_i) \rvo_i w_i}{\sum_{i=1}^n (1-\rva_i) w_i}
\end{equation} 
where $w_i$ are the so-called overlap weights given by 
\begin{equation}
	w_i  =
	\begin{cases}
		1-\hat e(\rvx_i) & \text{if $\rva_i = 1$} \\
		\hat e(\rvx_i) & \text{if $\rva_i = 0$.} 
	\end{cases}
	\label{eq:overlapwts}
\end{equation}
Notably, the overlap weights smoothly down-weigh subjects in the tails of the propensity score distribution, thereby mitigating the common problem of extreme propensity scores. In contrast, the standard inverse probability weights can suffer from excessive variance and correspond to emphasis on a target population which may consist of subjects very atypical for their particular value of $\rva$. 

In our experiments we considered two possibilities for $U_n(\theta)$. In the first, we measure the average effect in the overlap population of $\rva$ on the penultimate layer of the neural network, $\rvh^{(L-1)}$ leading to
\begin{equation}
	U_n(\theta) = \left \vert \hat \tau_{ATO}(\rvh^{(L-1)}) \right \rvert = \frac{\sum_{i=1}^n \rva_i \rvh_i^{(L-1)} w_i}{\sum_{i=1}^n \rva_i w_i} -  \frac{\sum_{i=1}^n (1-\rva_i) \rvh_i^{(L-1)} w_i}{\sum_{i=1}^n (1-\rva_i) w_i}
	\label{penultimate_layer}
\end{equation}
where $w_i$ are the overlap weights as in \eqref{eq:overlapwts}.
In the second option for $U_n$, we measure the average effect of $\rva$ on all intermediate layers of the neural network, leading to
\begin{equation}
	U_n(\theta) = \sum_{l=1}^{L-1} \left \lvert  \hat \tau_{ATO}(\rvh^{(l)}) \right \rvert.
	\label{all_layers}
\end{equation}
The benefits of learning fair internal representations in a neural network was recognized in \citet{Madras2018}. In particular, we may expect doing so can safeguard against bias in further downstream analyses, such as transfer learning.

\section{Experiments}
\label{sec:experiments}
We apply the proposed Pareto front estimation procedure to  two benchmarking datasets in the algorithmic fairness literature: the ProPublica recidivism dataset and the UCI adult income dataset.  The two datasets are summarized in Table \ref{table:datasets}. We also examine two possibilities for $U_n$ in Algorithm \ref{alg} corresponding to either \eqref{penultimate_layer} or \eqref{all_layers}, i.e., calculating the ATO in either the penultimate layer or all intermediate layers of the neural network respectively.

\begin{table*}[t]
	\caption{Dataset descriptions}
	\label{table:datasets}
	\centering
	\begin{tabular}{@{}rrrrrrr@{}}\toprule
		& \multicolumn{3}{c}{dataset features} \\
		\cmidrule{2-4} 
		Dataset & $dim(\rvx)$ & binary outcome $y$ & sensitive $\rva$ & training size & testing size & minibatch size \\ \midrule
		Recidivism & 12 & Re-offend in 2 years? & binary race  & 3086 &3086 & 150\\
		UCI & 93 & Income above 50K?  & binary race & 15470&15470 & 1000 \\
		UCI & 93 &  Income above 50K? & binary gender & 15470&15470 & 1000\\
		\bottomrule
	\end{tabular}
\end{table*}

\paragraph{Data preprocessing}
Missing values were preprocessed according to the accompanying code. In the UCI dataset, we wish to predict whether an individual has income above $50$K USD while remaining fair with respect to \textit{gender}. Separately, we wish to perform the same prediction task in the UCI dataset while remaining fair with respect to \textit{race}. In the recidivism dataset, we wish to predict whether an individual will recommit a crime in two years while remaining fair with respect to \textit{race}. In total we have three datasets  -- UCI (gender), UCI (race), and recidivism.

\paragraph{Comparison to alternatives}

For comparison, we implement, as a representative of the regularization approach, the adversarial learning technique proposed in \citet{Louppe2017}, which is not based on any specific fairness criterion. The idea is intuitive: the classifier and adversarial are engaged in a zero-sum game. Let $\theta_{clf}$ and $\theta_{adv}$ denote the parameters of the classifier and adversarial network, respectively. The classifier network attempts to make the best possible prediction of the binary output $\rvy$ given the input $\rvx$, while ensuring that $\rva$ cannot be predicted well from the predicted score $\hat p$.  The adversary, on the other hand, attempts to make the best possible prediction of the binary sensitive attribute $\rva$ based on the classifier's prediction $\hat p$. The adversary behaves according to the objective,
$$
\min_{\theta_{adv}} Loss_{\rva}(\theta_{clf},\theta_{adv}),
$$
where $Loss_{\rva}$ is associated to the prediction of the sensitive attribute $\rva$ given $\hat p$. The classifier behaves according to
\begin{equation}
	\argmin_{\theta_{clf}} \left [ 	Loss_{\rvy}(\theta_{clf}) - \beta Loss_{\rva}(\theta_{clf},\theta_{adv})	\right ].
	\label{eq:adv}
\end{equation}
where $\beta \in [0,1]$ and $Loss_{\rvy}$ measures the loss of predicting $\rvy$ based on $\rvx$. The binary cross-entropy loss is employed in both objectives.

\subsection{Experimental details}

For each of the three datasets -- UCI (gender), UCI (race), recidivism -- the data is split into 100 training-testing sets, with the split reported in Table \ref{table:datasets}. Then, for each of the 100 training sets, we apply Algorithm 1 to find Pareto front candidates with $R_n$ corresponding to the binary cross-entropy loss, $U_n$ given by either \eqref{penultimate_layer} or \eqref{all_layers}, and $\Lambda$ containing fifteen different $\lambda$ values in the interval $[0,1]$, approximately evenly-spaced on the log scale. Exact values used for $\Lambda$ can be consulted in the accompanying code. 

\paragraph{Propensity score network} 
To calculate $U_n$ in either \eqref{penultimate_layer} or \eqref{all_layers}, propensity scores must first be estimated. We used the same neural network architecture in all three datasets to estimate the propensity scores $P(\rva=1 \mid \rvx)$. The network has three fully-connected layers, with 32 hidden units each, interspersed with a dropout layer with dropout probability $0.2$. The ReLU activation function is used in all intermediate layers while the sigmoid function is used in the output layer.

Since the propensity score network is performing binary classification of $\rva$ based on input $\rvx$, we used the binary cross-entropy loss. 
The Adam optimisation algorithm \citep{Kingma2014} was used to train the propensity network. The learning rate is set to $0.001$. Training took place over 100 epochs. Mini-batch size is reported in Table \ref{table:datasets} and was chosen to be around $5\%$ of the training set size.  After the propensity  network is trained, we  calibrate the probability prediction according to the methodology proposed in \citet{Guo2017} where we used their GitHub code with no modification. 

\paragraph{Architecture $f_\theta$}
The architecture of the neural network $f_\theta$ in Algorithm \ref{alg} is reported in Table \ref{table:architecture}. The number of fully-connected layers and number of hidden nodes in each layer (held constant over the layers) were tuned for each data setting with the goal of not incurring over-fitting in the held-out test set. Each fully-connected layer is interspersed with a dropout layer with dropout probability $0.2$. The ReLU activation function is used in all intermediate layers while the sigmoid function is used in the output layer.

\begin{table}[h]
	\caption{$f_\theta$ network architecture}
	\label{table:architecture}
	\center
		\begin{tabular}{@{}rrrr@{}}\toprule
			& \multicolumn{2}{c}{neural network features} \\
			\cmidrule{2-3} 
			Dataset & layers $L$ & hidden nodes \\ \midrule
			Recidivism & 4 & 4  \\
			UCI & 32 & 10  \\
			UCI & 32 & 10  \\
			\bottomrule
		\end{tabular}
\end{table}

To learn the network $f_\theta$ in Algorithm \ref{alg}, we again use Adam. The initial learning rate is set to $0.001$. We reduce the learning rate when the training loss has stopped decreasing by using the \texttt{ReduceLROnPlateau} scheduler in PyTorch, setting the factor and patience variables to $0.9$ and $10$, respectively. 
All training took place over 500 epochs. Mini-batch sizes are as reported in Table \ref{table:datasets}.

\paragraph{Adversarial alternative}
Our implementation of the adversarial technique of \citet{Louppe2017} is based on GoDataDriven's code base. The following steps are alternated over 200 epochs: (1) train the adversarial network for a \textit{single epoch}, holding the classifier network fixed and (2) train the classifier network on a \textit{single sampled mini batch}, holding the adversarial network fixed.

The adversarial network has 4 hidden layers with 32 hidden units in each. The ReLU activation was used throughout except in the final layer where the sigmoid function is used. The adversarial network was pretrained for 5 epochs. 
For the classifier network, we employed the same network as that of $f_\theta$ in Algorithm \ref{alg} and kept all training choices, such as the optimisation algorithm and mini-batch size, the same. The classifier was pretrained for 2 epochs. Further implementation details for the adversarial approach can be found in the accompanying code.

\begin{table}[h]
	\caption{In visualizing the estimated Pareto front, we plot for each candidate $\theta$ its value in the fairness-accuracy space $R(\theta;\Ptest)$ versus $U(\theta;\Ptest)$, as given below for four different choices of $U(\theta;\Ptest)$. The development of the $MV$ estimators are introduced in the supplementary material.}
	\label{table:unfairness_measures}
	\vskip 0.15in
	\begin{center}
		\begin{sc}
			\begin{tabular}{lr}
				\toprule
				ATO & $ \left \lvert  \tau_{ATO}(f_\theta(\rvx);\Ptest) \right \rvert $ \\
				\hline
				Equal odds (EO) &  $MV(f_\theta(\rvx),\rva \mid \rvy; \Ptest)$ \\
				\hline
				Equal opportunity (EOpp) & $MV(f_\theta(\rvx),\rva \mid \rvy=1; \Ptest)$ \\
				\hline
				Demographic parity (DP) & $ MV(f_\theta(\rvx),\rva; \Ptest)$ \\
				\bottomrule 
			\end{tabular} 
		\end{sc}
	\end{center}
	\vskip -0.1in
\end{table}

\subsection{Results}
\label{results}
In summary, we have three datasets and apply Algorithm \ref{alg} with two possibilities of $U_n$. We also apply the adversarial technique to each of the three datasets for comparison. 

\begin{figure}[h]
	\includegraphics[width=0.47\textwidth]{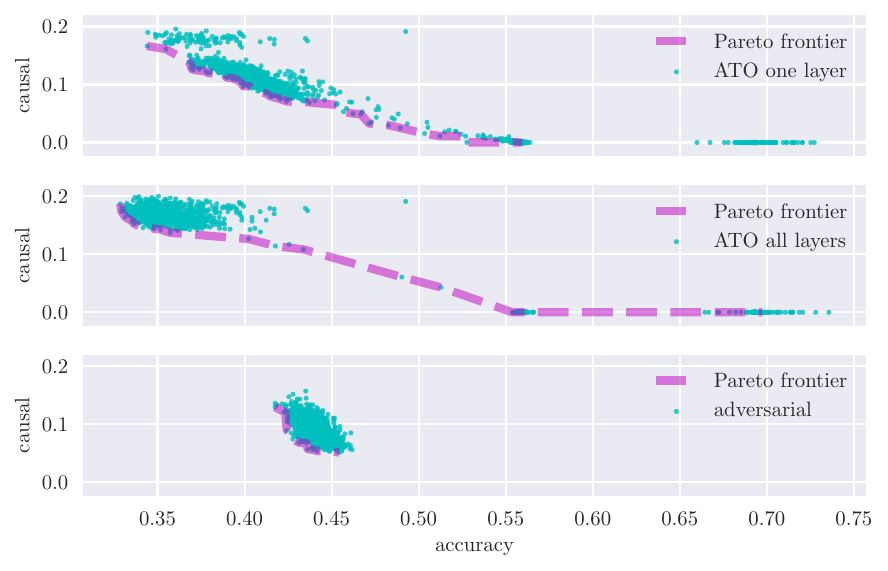}
	\includegraphics[width=0.47\textwidth]{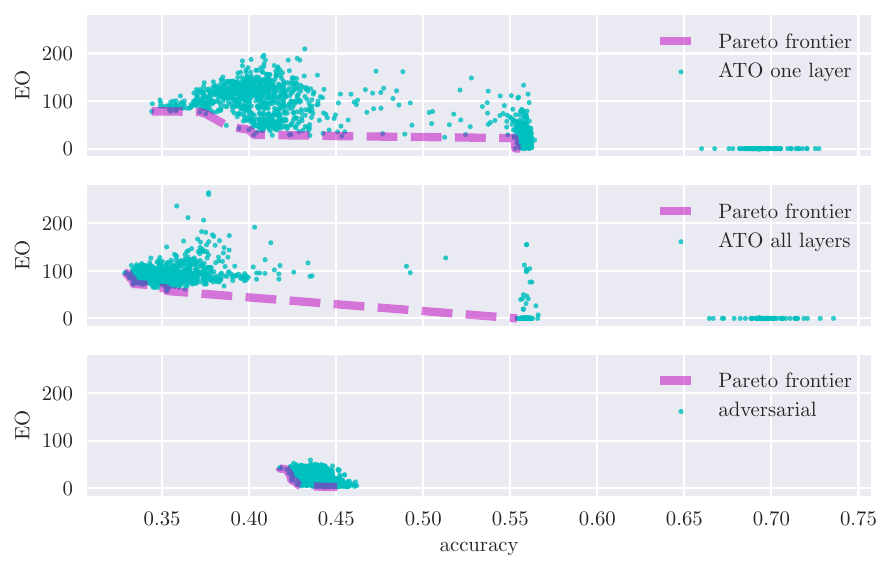}
	\includegraphics[width=0.47\textwidth]{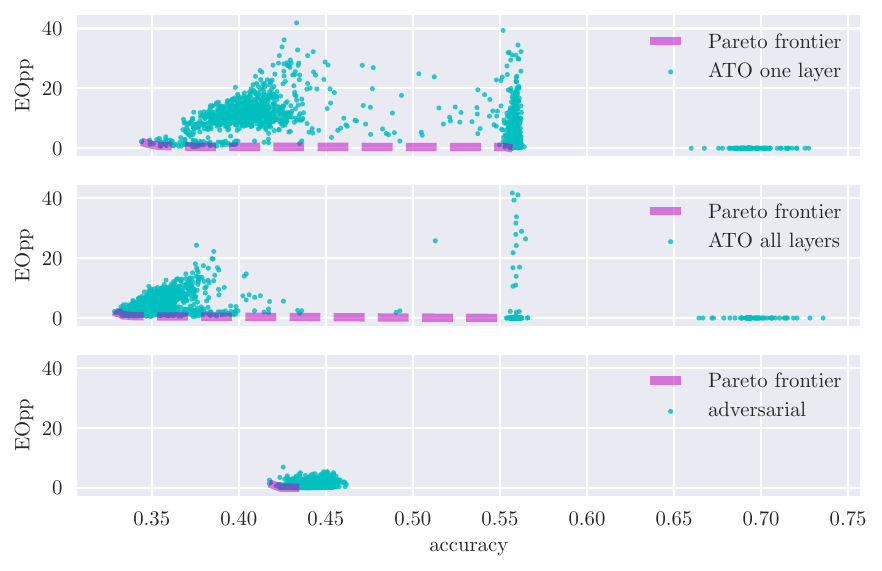} \quad \quad \quad
	\includegraphics[width=0.47\textwidth]{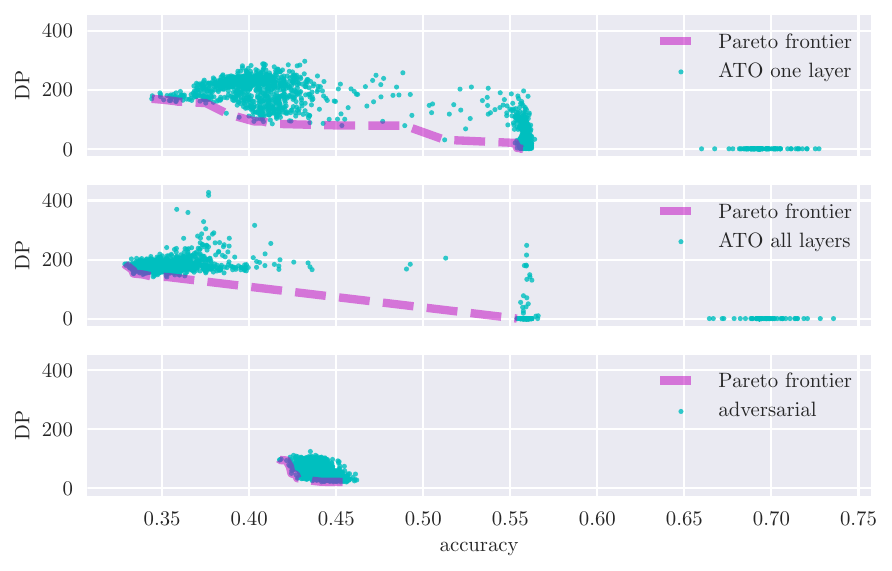}
	\caption{Pareto front estimation in the UCI (gender) test set. Each three by one block corresponds to a different fairness evaluation in Table \ref{table:unfairness_measures}. Within each block, the row corresponds to different Pareto front estimation procedures  -- ``ATO one layer" refers to Algorithm \ref{alg}  employing the ATO causal estimand in the penultimate layer as $U_n$, ``ATO all layers" refers to Algorithm \ref{alg} with the ATO causal estimand calculated over all intermediate layers, and lastly the adversarial approach. In each subplot, 1500 candidates are displayed along with the culled Pareto front in dashed magenta.  The accuracy x-axis plots $R( \theta,\hat{P}_{\rm{test}})$ and the fairness y-axis plots $U( \theta,\hat{P}_{\rm{test}})$.
		\label{pareto_UCIgender}}
\end{figure}

We present the results for the UCI (gender) dataset in Figure~\ref{pareto_UCIgender}. The results for UCI (race) and recidivism can be found in Figure~\ref{pareto_UCIrace} and Figure~\ref{pareto_recidivism} in Appendix \ref{app:add_figures}, respectively. 
Each three-by-one block in Figure \ref{pareto_UCIgender} corresponds to a different fairness measure $U(\theta,\Ptest)$ in Table \ref{table:unfairness_measures}. Within each block, the rows correspond to a different approach to finding the Pareto candidates. 

Since 15 values of $\lambda$ and 100 training-testing splits are considered, the output of Algorithm \ref{alg}, \textsc{candidates}, consists in total 1500 learned network parameters. The 1500 experiments were run in parallel using high performance computing resources. It should be noted that the individual nodes used are less powerful than many personal workstations, and even so, a single training run takes only about thirty minutes. 

The results of applying Algorithm \ref{alg} with $U_n(\theta) = \left \vert \hat \tau_{ATO}(\rvh^{(L-1)}) \right \rvert$  and $U_n(\theta) = \sum_{l=1}^{L-1} \left \lvert  \hat \tau_{ATO}(\rvh^{(l)}) \right \rvert$ are shown in the first and second rows of each three-by-one block in Figure \ref{pareto_UCIgender}, respectively. In each sub-figure, we plot for each $\theta \in \textsc{candidates}$, its value in the fairness-accuracy space, $R( \theta,\Ptest)$ versus $U( \theta,\Ptest)$. We further display the Pareto front culled from these 1500 Pareto candidates where the culling simply discards the dominated points.

The result of the adversarial approach is displayed in the third row of each three-by-one block of Figure \ref{pareto_UCIgender}. Fifteen values of $\lambda$ in \eqref{eq:adv} along with 100 training-testing splits are considered, producing in total 1500 $\theta_{clf}$, each of which is plotted in the fairness-accuracy space, $R( \theta,\Ptest)$ versus $U( \theta,\Ptest)$.


We can immediately see from Figure \ref{pareto_UCIgender} that compared to Algorithm \ref{alg}, the adversarial approach is less capable of finding a Pareto front that spans the fairness-accuracy space. Indeed Table \ref{table:dominatedpointscount} shows that the set of non-dominated points  found by the adversarial approach is much smaller relative to Algorithm \ref{alg}.

We also observe better Pareto front estimation when Algorithm \ref{alg} is applied with $U_n$ as the ATO causal measure calculated in the penultimate layer, compared to the ATO calculated over all layers. This seems to be true of the two other data settings as well, as can be seen in Appendix \ref{app:add_figures}. This suggests using \eqref{all_layers} for $U_n$ makes training the neural network more difficult. It may be worthwhile to explore training the network one layer at a time in future work.

Finally, from Figure \ref{pareto_UCIgender}, we can see that the demographic parity, equal odds, and equal opportunity fairness measures tend to form distinct clumps in the fairness-accuracy space, relative to the ATO causal measure.
Results for UCI (race) and recidivism contained in Appendix \ref{app:add_figures} indicate much of the same conclusions.
Appendix \ref{app:add_figures} also contains further visualization on the distributions of the prediction probabilities as $\lambda$ in Algorithm \ref{alg} is dialed between 0 and 1.

\begin{table}[h!]
	\caption{
		Each row corresponds to a different approach of finding the fairness-accuracy Pareto front. Each column corresponds to a different dataset. The individual cells report the number of \textit{non-dominated} points (higher is better) calculated in the space $(R(\theta,\Ptest), U(\theta,\Ptest))$ where $U(\theta,\Ptest) =  \left \lvert  \tau_{ATO}(f_\theta(\rvx);\Ptest) \right \rvert $. ATO one layer refers to Algorithm \ref{alg} where $U_n$ is the ATO calculated on the penultimate layer, while ATO all layers refers to the ATO calculated over all intermediate layers.
	}
	\label{table:dominatedpointscount}
	\begin{center}
		\begin{tabular}{|c|c|c|c|}
			\hline 
			& UCI (gender) & UCI (race) & Recidivism  \\ 
			\hline 
			ATO one layer & 44 & 33 & 89 \\ 
			\hline 
			ATO all layers & 27 & 27 & 53 \\ 
			\hline 
			Adversarial & 13 & 9 & 27 \\ 
			\hline 
		\end{tabular} 
	\end{center}
\end{table}

\section{Future work}

We have found standardization of the objectives to be crucial as a preprocessing step. Recent works such as GradNorm \citep{chen_gradnorm_2018} may offer a way to simultaneously standardize as we train rather than as a preprocessing step. 
We also discovered that the Chebyshev scalarization scheme, though superior to the linear scalarization scheme, is still seen to often produce dominated points.
This may not be a particular fault of the Chebyshev scalarization scheme, Rather the phenomenon is likely due to the fact that we can never perform optimization perfectly, especially for deep neural networks using stochastic optimization algorithms. 
Barring this practical difficulty, as Chebyshev scalarization only leads to weakly Pareto optimal solutions, there still may be theoretical grounds for improvement. We may do better by using e.g., the hypervolume indicator \citep{auger_hypervolume-based_2012} to find points that are Pareto optimal. This has its own challenges, as the hypervolume indicator is expensive to calculate in higher dimensions. 

Our experiments also reveal the difficulty of selecting weights in the Chebyshev scalarization scheme in such a way that would lead to a wide diversity of Pareto optimal points. In fact, many different weights often produced nearly identical solutions. Appropriately specifying the weights in a scalarization scheme without a priori information on the shape of the Pareto front itself is an open problem in multi-objective optimization. We had stated earlier that adaptive weights would be an improvement to the current methodology. This certainly merits further study but the following challenges can be anticipated. Adapting the weights while training the neural network fundamentally changes the convergence behavior of the optimization algorithm since the optimization problem itself is gradually changing. It is already a nontrivial affair to train a deep neural network, and so we expect the addition of weight adaptation will require careful tinkering to the training process. Furthermore, many adaptive weight schemes seem to depend on the Pareto front shape itself. Without a better understanding of the Pareto front, we might do more harm by using an unsuitable weight adaptation technique.

	\bibliography{references}
	\bibliographystyle{plainnat}

	\newpage
	\appendix
\appendix
\section{Additional figures}
\label{app:add_figures}
\begin{figure}[h!]
	\begin{center}
		\includegraphics[width=0.47\textwidth]{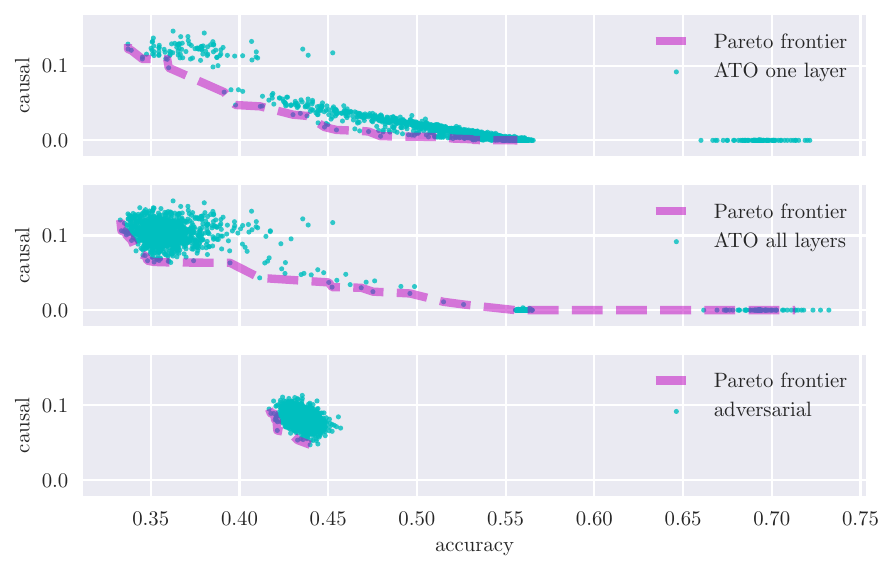}
		\includegraphics[width=0.47\textwidth]{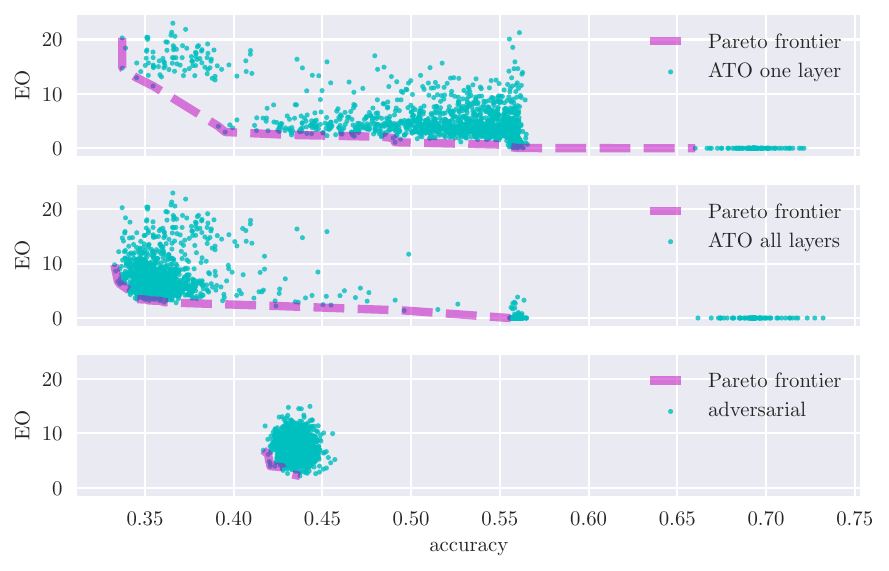}
		\includegraphics[width=0.47\textwidth]{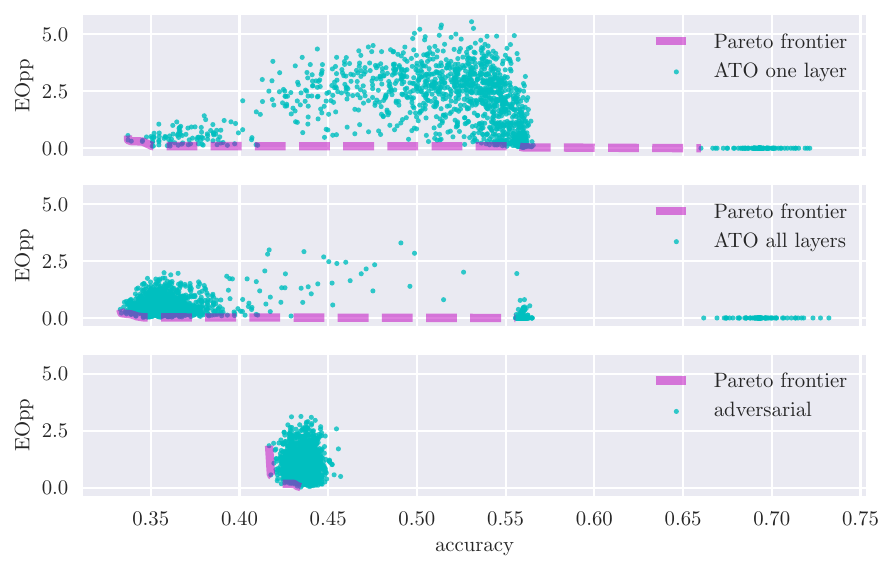}
		\includegraphics[width=0.47\textwidth]{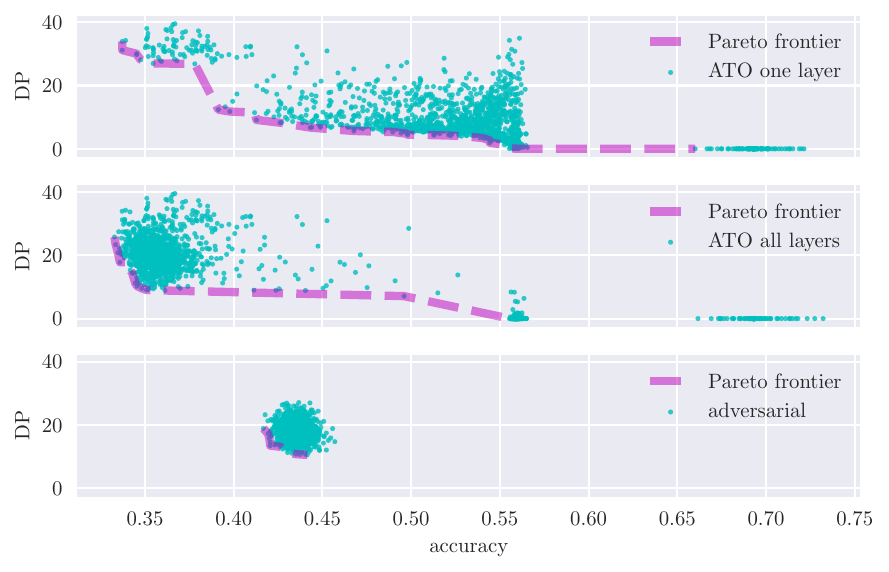}
		\caption{Pareto front estimation in the UCI (race) data set. Each three by one block corresponds to a fairness measure in Table \ref{table:unfairness_measures} of the main document. Within each block, the row corresponds to different Pareto front estimation  procedure -- ``ATO one layer" refers to Algorithm \ref{alg} in the main document employing the ATO causal estimand in the penultimate layer as $U_n$, ``ATO all layers" refers to Algorithm \ref{alg} with the ATO causal estimand calculated over all layers, and lastly the adversarial approach. In each subplot, 1500 candidates are displayed along with the culled Pareto front displayed in dashed magenta.  
		}
		\label{pareto_UCIrace}
	\end{center}
	\vskip -0.3in
\end{figure}

\begin{figure}[h!]
	\begin{center}
		\includegraphics[width=0.47\textwidth]{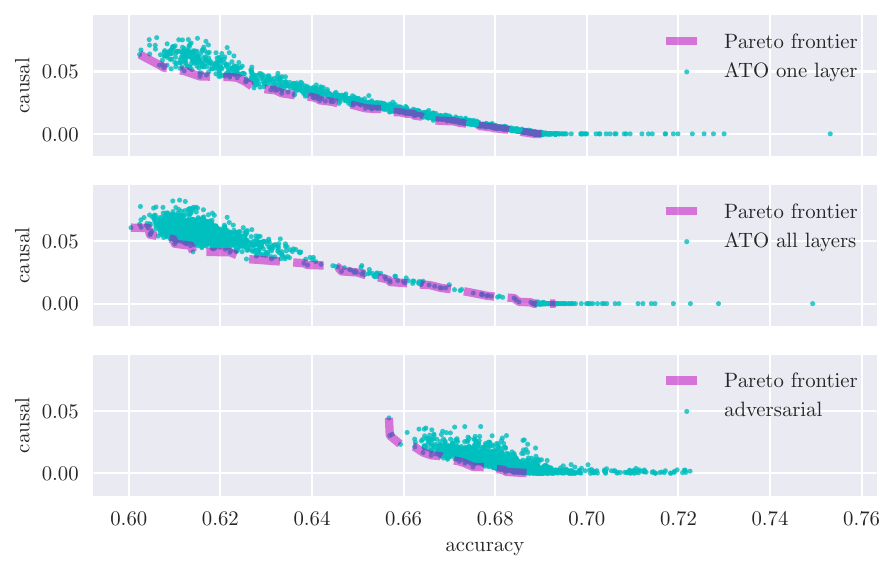}
		\includegraphics[width=0.47\textwidth]{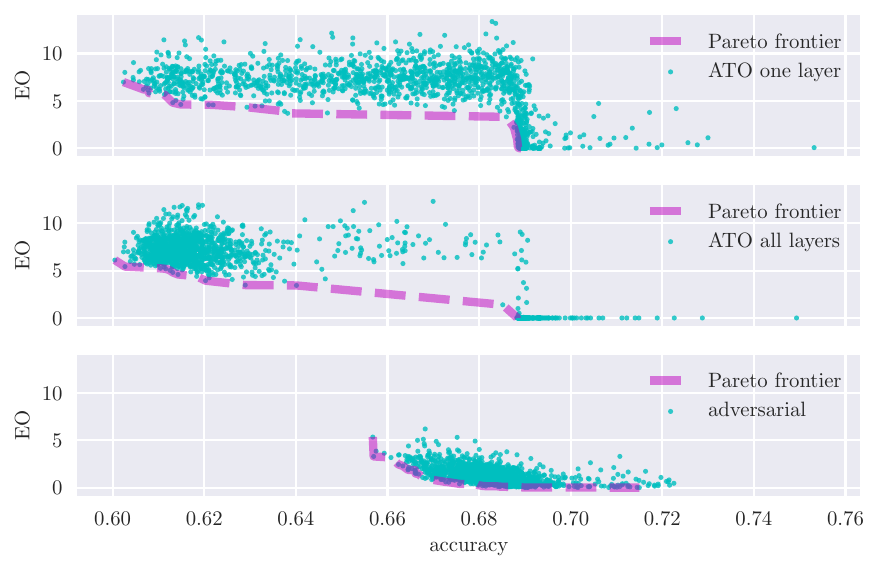}
		\includegraphics[width=0.47\textwidth]{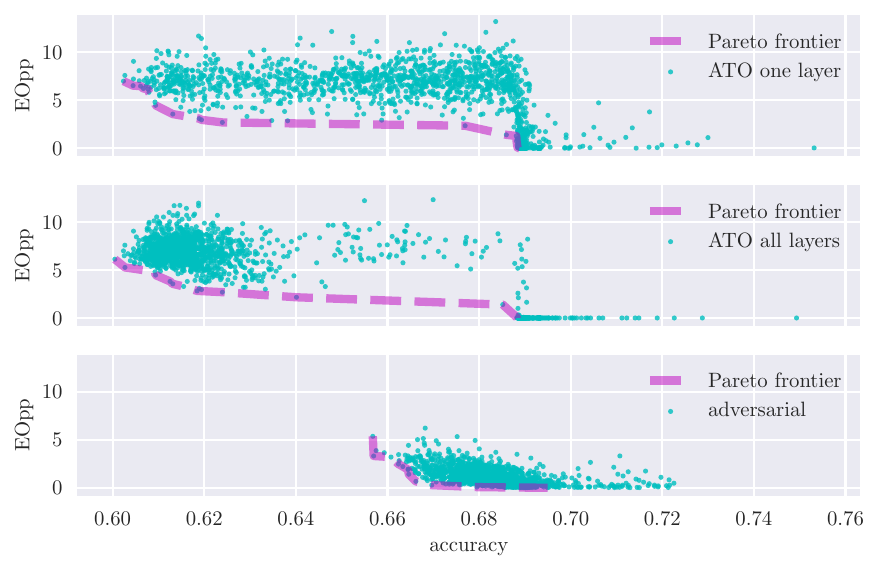}
		\includegraphics[width=0.47\textwidth]{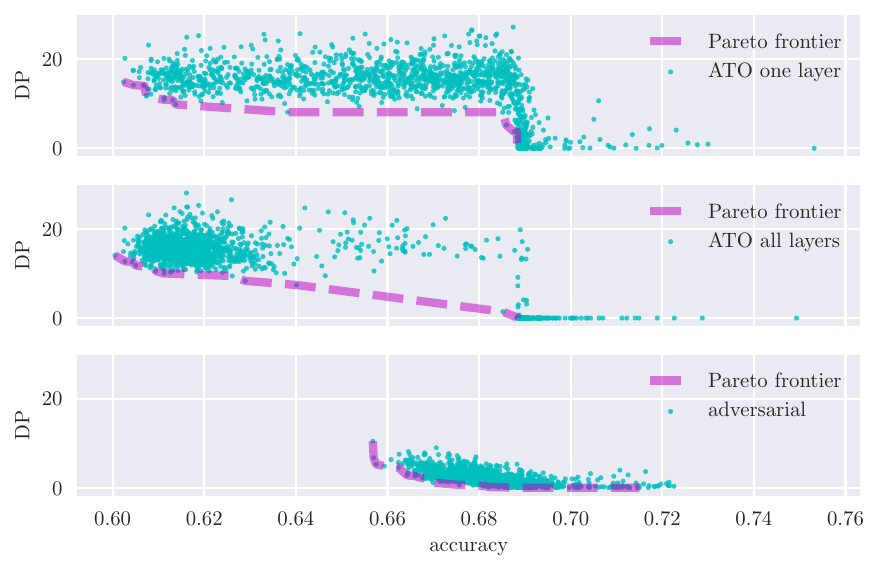}
		\caption{Pareto front estimation in the recidivism data set. Each three by one block corresponds to a fairness measure in Table \ref{table:unfairness_measures} of the main document. Within each block, the row corresponds to different Pareto front estimation  procedure -- ``ATO one layer" refers to Algorithm \ref{alg} in the main document  employing the ATO causal estimand in the penultimate layer as $U_n$, ``ATO all layers" refers to Algorithm \ref{alg} with the ATO causal estimand calculated over all layers, and lastly the adversarial approach. In each subplot, 1500 candidates are displayed along with the culled Pareto front displayed in dashed magenta.  
		}
		\label{pareto_recidivism}
	\end{center}
	\vskip -0.3in
\end{figure}

In this section, we provide additional figures for the experiments conducted in ``The Fairness-Accuracy Pareto Front.''  Figures \ref{pareto_UCIrace} and \ref{pareto_recidivism} here are analogous to Figure \ref{pareto_UCIgender} in the main text. The figures report, respectively, the results for the UCI (race) and recidivism dataset. Within each three-by-one block, the results of various Pareto estimation procedures are reported. Each block of subfigures corresponds to a different $U(\theta;\Ptest)$.

\begin{figure}[h!]
	
	\includegraphics[width=0.49\textwidth]{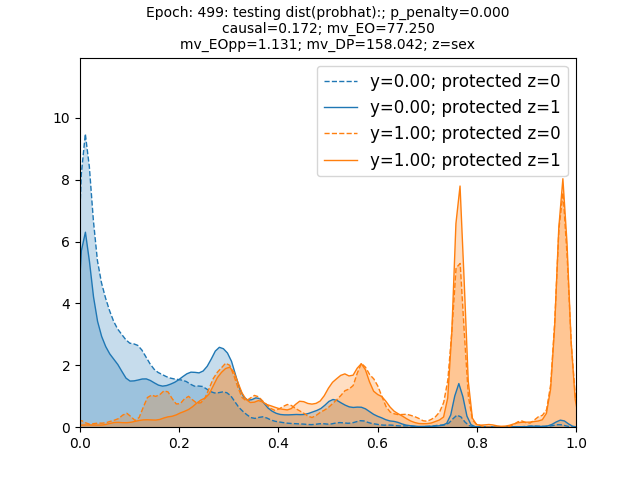}
	\includegraphics[width=0.49\textwidth]{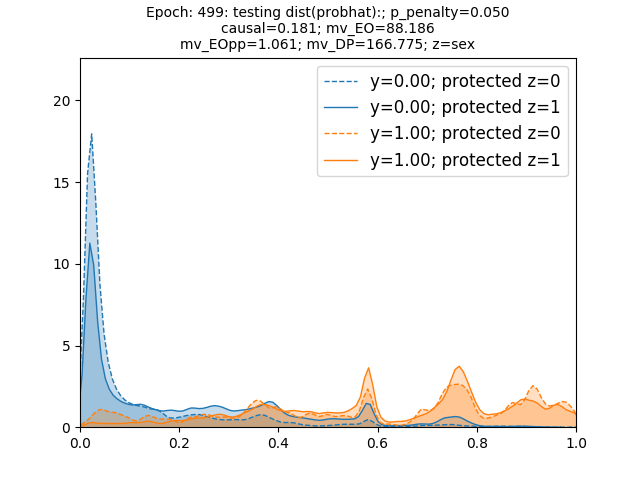}
	\includegraphics[width=0.49\textwidth]{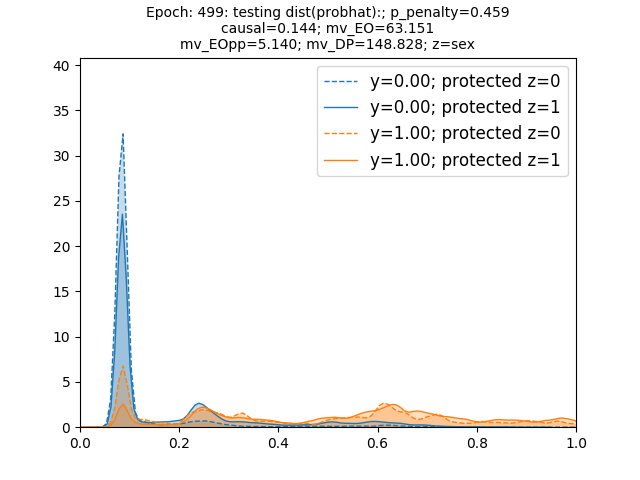}
	\includegraphics[width=0.49\textwidth]{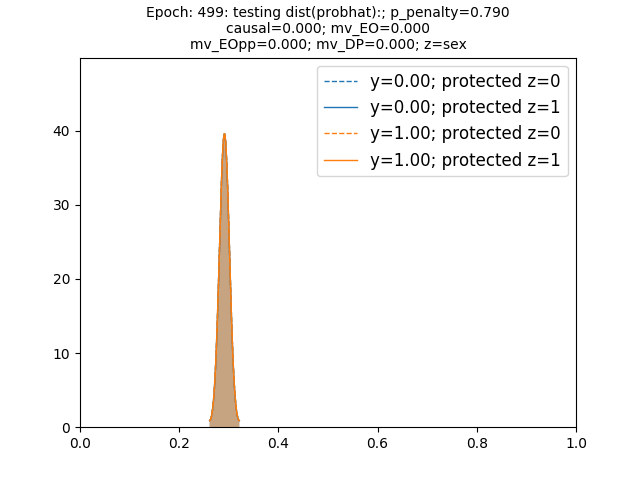}		
	\caption{For a particular training-testing split of the UCI (gender) dataset, we display the distributions of the predicted probabilities in the test set for four different values of $\lambda$, where $\lambda$ is indicated by \textsc{p\_penalty} in the heading of each plot. The distributions are broken down by different values of the true target label $\rvy$ and the sensitive attribute $\rvz$. 
		Besides the ATO measure (\textsc{causal}), we also indicate equalised odds (\textsc{mv\_EO}), equality of opportunity (\textsc{mv\_EOpp}) and demographic parity (\textsc{mv\_DP}) in the headings of the subplots. 		\label{fig:probdist_UCIgender}}	
\end{figure}

\begin{figure}[h!]
	\includegraphics[width=0.49\textwidth]{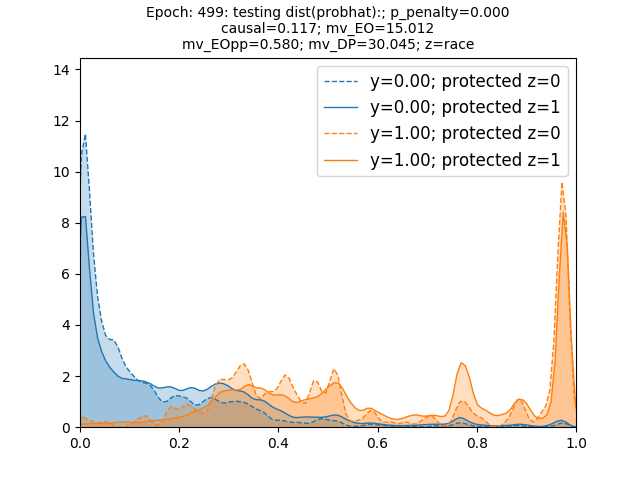}
	\includegraphics[width=0.49\textwidth]{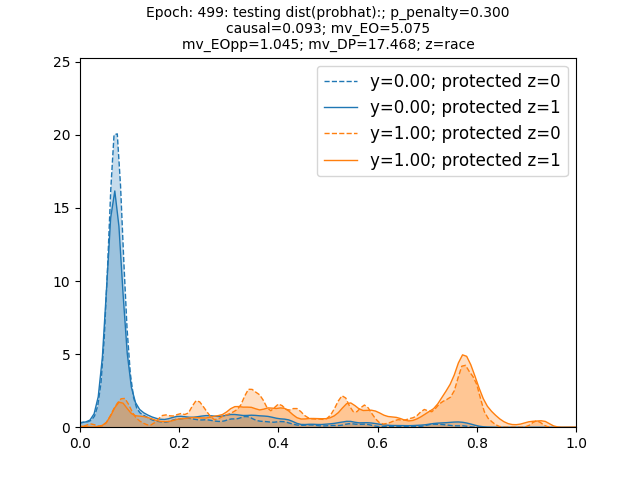}
	\includegraphics[width=0.49\textwidth]{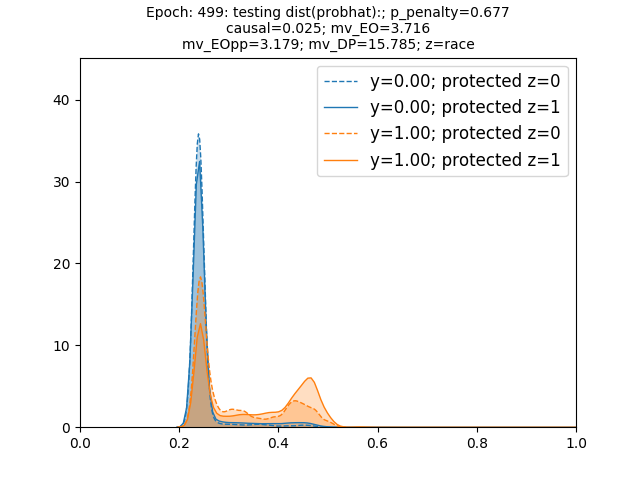}
	\includegraphics[width=0.49\textwidth]{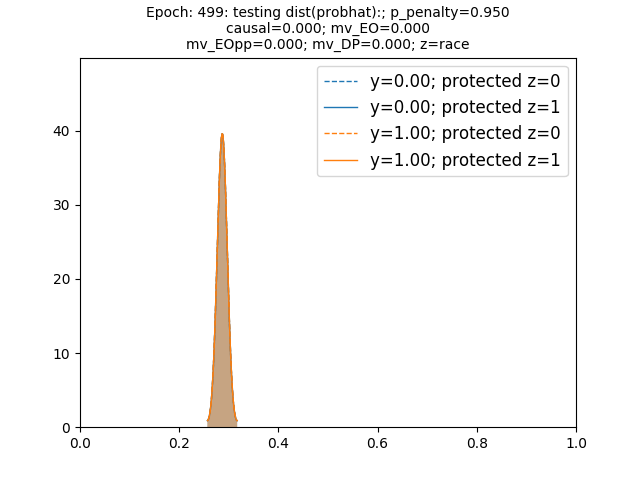}		
	\caption{For a particular training-testing split of the UCI (race) dataset, we display the distributions of the predicted probabilities in the test set for four different values of $\lambda$, where $\lambda$ is indicated by \textsc{p\_penalty} in the heading of each plot. The distributions are broken down by different values of the true target label $\rvy$ and the sensitive attribute $\rvz$. 
		Besides the ATO measure (\textsc{causal}), we also indicate equalised odds (\textsc{mv\_EO}), equality of opportunity (\textsc{mv\_EOpp}) and demographic parity (\textsc{mv\_DP}) in the headings of the subplots.		\label{fig:probdist_UCIrace}}
\end{figure}

\begin{figure}[h!]
	
	\includegraphics[width=0.49\textwidth]{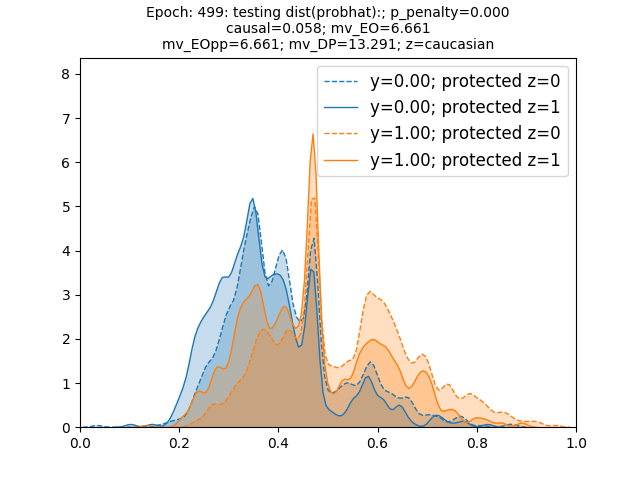}
	\includegraphics[width=0.49\textwidth]{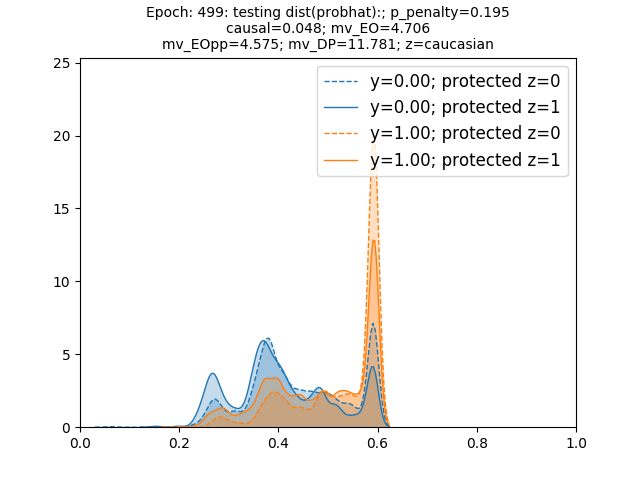}
	\includegraphics[width=0.49\textwidth]{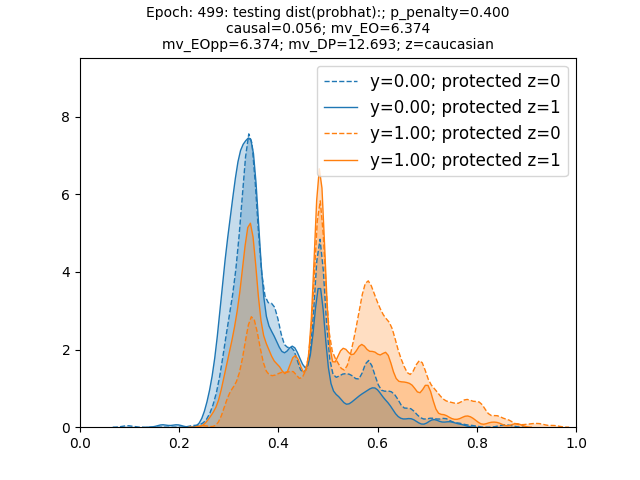}
	\includegraphics[width=0.49\textwidth]{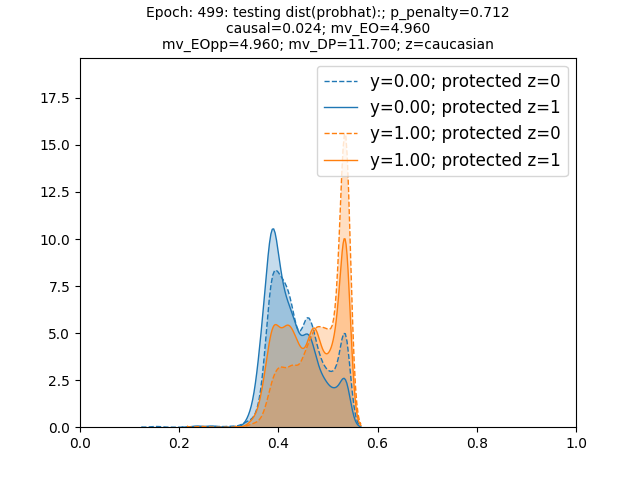}		
	\caption{For a particular training-testing split of the recidivism dataset, we display the distributions of the predicted probabilities in the test set for four different values of $\lambda$, where $\lambda$ is indicated by \textsc{p\_penalty} in the heading of each plot. The distributions are broken down by different values of the true target label $\rvy$ and the sensitive attribute $\rvz$. 
		Besides the ATO measure (\textsc{causal}), we also indicate equalised odds (\textsc{mv\_EO}), equality of opportunity (\textsc{mv\_EOpp}) and demographic parity (\textsc{mv\_DP}) in the headings of the subplots.		\label{fig:probdist_recidivism}}
\end{figure}

We also provide figures that help visualise the effect of dialling $\lambda$ in Algorithm \ref{alg} of the main text from 0 to 1.  
In Figure  \ref{fig:probdist_UCIgender}, we display the distribution of the classifier's prediction in the UCI (gender) dataset broken down by class label and sensitive attribute. Each panel of Figure \ref{fig:probdist_UCIgender} is a different $\lambda$ value. In addition to reporting the ATO measure of fairness, we also indicate other non-causal fairness metrics including Equalised Odds, Equal Opportunity, and Demographic Parity. 
Similar visualisation for the UCI (race) and recidivism dataset can be found in Figures \ref{fig:probdist_UCIrace} and \ref{fig:probdist_recidivism}, respectively. 

\section{Estimation of demographic and conditional parity}
We describe a quantitative index that measures the degree to which conditional parity holds. 
First, recall we say a prediction score $\hat S$ exhibits conditional parity with respect to sensitive attribute $Z$ conditioned on $U$ if $Z$ and $U$ are independent conditional on $U$. We will limit the scope to categorical $U$ and categorical $Z$. The conditional independence statement $\hat S \independent Z \mid U$ is typically assessed using the Cochran-Mantel-Haenszel test when $\hat S$ and $Z$ are binary variables and $U$ is another, let's say $k$-level, categorical variable. Conditional independence in this case simply reduces to equality of odds ratios in each of the $k$ two-by-two contingency tables. 

A more sophisticated method is required to assess conditional parity when $\hat S$ is continuous. In the following, we introduce an index for conditional parity by adapting the mean-variance statistic of ~\cite{Cui2015ModelFreeFS}, which is not directly applicable as it is designed for \textit{unconditional} independence assessment between a continuous variable and a categorical variable. 

\subsection{The Mean-Variance Index}
Without loss of generality, suppose the categorical random vector $Z$ takes value in the set $\{z_1,\ldots,z_R\}$. Let $F_r(s) = P(\hat S \le s \mid Z=z_r)$ be the conditional distribution of $\hat S$ given $Z=z_r$. Let $F(s) = P(\hat S \le s)$ be the unconditional distribution of $\hat S$. The mean-variance index \citet{Cui2015ModelFreeFS} is given by
\begin{equation}
	MV = \sum_{r=1}^R P(Z=z_r) \int [F_r(s) - F(s) ]^2 \,dF(s),
	\label{MV_uncond_pop}
\end{equation}
where $P(Z=z_r)>0$ for all $j =1, \ldots,r$.  Note that the integral in \eqref{MV_uncond_pop} is simply the Cram\'er-von Mises distances between $F_r$ and $F$, and so the mean-variance statistic is the weighted average of these distances, weighted by how likely a particular value of $Z$ is. If $\hat S$ and $Z$ are independent, then $F_r(s) = F(s)$ for all $r=1,\ldots,R$. Thus the mean-variance index in  \eqref{MV_uncond_pop} has the salient property that it is zero if and only if $\hat S \independent Z$. 

Let the notation $1\{\cdot\}$ denote indicator function of an event, i.e.\ it is 1 if the event happens and 0 otherwise. The plug-in estimator for \eqref{MV_uncond_pop} based on a sample $\{(\hat S_i,Z_i), i=1,\ldots,n\}$ is 
\begin{equation}
	\widehat{MV} =  \sum_{r=1}^R \hat P(Z=z_r) \left[ \frac{1}{n}\sum_{i=1}^n  [\hat F_r(\hat S_i)	- \hat F(\hat S_i)	]^2 \right]
	\label{MV_uncond_sample}
\end{equation}
where 
\begin{align*}
	\hat P(Z=z_r) &= n^{-1} \sum_{i=1}^n 1\{Z_i = z_r\}, \\
	\hat F_r(s)&= \frac{\sum_{i=1}^n 1\{\hat S_i \le s, Z_i = z_r\}}{\sum_{i=1}^n 1\{Z_i = z_r\}}, \\
	\hat F(s) &= \frac{1}{n} \sum_{i=1}^n 1\{\hat S_i \le s\}.
\end{align*}
That the estimator in \eqref{MV_uncond_sample} is consistent for its theoretical counterpart in  \eqref{MV_uncond_pop},  i.e.\ $\widehat{MV}$ converges in probability to $MV$ as $n \to \infty$, is established in \citet{Cui2015ModelFreeFS}. 

\subsection{A new index to measure conditional parity}
\label{appendix:mv_condpar}

Now let us return to the assessment of $\hat S \independent Z \mid U$. Suppose the categorical random vector $U$ takes value in the set $\{u_1,\ldots,u_K\}$. Then we propose the following as a direct extension of the mean-variance statistic of \citet{Cui2015ModelFreeFS} to the conditional case:
\begin{equation}
	MV^k= \sum_{r=1}^R P_k(Z=z_r) \int [F_{r,k}(s) - F_k(s) ]^2 \,dF_k(s)
	\label{MV_condk_pop}
\end{equation}
where 
\begin{align*}
	P_k(Z=z_r) &= P(Z=z_r \mid U=u_k) \\
	F_{r,k}(s) &= P( \hat S \le s \mid Z = z_r, U = u_k) \\
	F_k(s) &= P( \hat S \le s \mid U = u_k ).
\end{align*}

We employ the following statistic to assess the degree to which $\hat S \independent Z \mid U$ holds:
\begin{equation}
	\max_{k = 1,\ldots,K} MV^k(\hat S, Z).
	\label{MV_cond_pop}
\end{equation}
Since $MV^k$ is zero if and only if $\hat S$ is independent of $Z$ conditioned on $U$ and the same is true for the maximum. In other words, \eqref{MV_cond_pop} is zero if and only if $\hat S \independent Z \mid U$.

Since \eqref{MV_condk_pop} and \eqref{MV_cond_pop} are both unknown population quantities, they require estimation. Let $\widehat{MV_k} $ be the plug-in estimator for \eqref{MV_condk_pop} based on the empirical measure. Namely, let
\begin{equation*}
	\widehat{MV^k}(\hat S, Z) = \sum_{r=1}^R \hat P_k(Z=z_r) \int [\hat F_{r,k}(s) - \hat F_k(s) ]^2 \,d \hat F_k(s)
\end{equation*}
where
\begin{align*}
	\hat P_k(Z=z_r) &=  \frac{\sum_{i=1}^n 1\{Z_i = z_r, U_i = u_k\}}{\sum_{i=1}^n 1\{U_i = u_k\}} \\
	\hat F_{r,k}(s) &= \frac{\sum_{i=1}^n 1\{\hat S_i \le s, Z_i  = z_r, U_i = u_k\}}{\sum_{i=1}^n 1\{Z_i = z_r, U_i = u_k\}} \\
	\hat F_k(s) &= \frac{\sum_{i=1}^n 1\{\hat S_i \le s, U_i = u_k\}}{\sum_{i=1}^n 1\{ U_i = u_k\}}.
\end{align*}

To estimate \eqref{MV_cond_pop}, we simply take the maximum over $k$:
\begin{equation}
	\max_{k = 1,\ldots,K} \widehat{MV^k}.
	\label{MV_cond_sample}
\end{equation}
We will call the quantity in \eqref{MV_cond_sample} the \textbf{conditional mean-variance index}.
Note that this measure will be zero if conditional independence holds and will increase with increasing dependence. Hence, it is a suitable measure to assess conditional parity.

\end{document}